# Optimizing the Parameters of A Physical Exercise Dose-Response Model: An Algorithmic Comparison

Mark Connor[a], Michael O'Neill[a]

[a] Natural Computing Research And Applications Group, Smurfit School of Business, University College Dublin, Ireland

**ABSTRACT**

The purpose of this research was to compare the robustness and performance of a local and global optimization algorithm when given the task of fitting the parameters of a common non-linear dose-response model utilized in the field of exercise physiology. Traditionally the parameters of dose-response models have been fit using a non-linear least-squares procedure in combination with local optimization algorithms. However, these algorithms have demonstrated limitations in their ability to converge on a globally optimal solution. This research purposes the use of an evolutionary computation based algorithm as an alternative method to fit a nonlinear dose-response model. The results of our comparison over 1000 experimental runs demonstrate the superior performance of the evolutionary computation based algorithm to consistently achieve a stronger model fit and holdout performance in comparison to the local search algorithm. This initial research would suggest that global evolutionary computation based optimization algorithms may present a fast and robust alternative to local algorithms when fitting the parameters of non-linear dose-response models.

## 1. Introduction

In has been observed that many biological systems exhibit nonlinear characteristics by examining the relationship between an input to the system, typically known as a dose, and the subsequent system output or response. In the context of a dose-response model related to physical exercise we consider the inputs to the system to be a combination of the volume and intensity of the exercise activity conducted, while the response is the adaption of the human body typically quantified by changes in responses measured over a battery of tests. A common example of this relationship is the cardiac adaptions that occur in response to an increase in the volume of an aerobic exercise activity such as running, swimming or cycling.

A traditional approach to fitting dose-response models is to use non-linear least squares regression (NLS) [10]. Although there has been a long history of using NLS to fit the parameters of dose-response models, the algorithms predominately used to perform the parameter search during the fitting process have been from the family of local optimizers. These algorithms take iterative steps towards a minimum value of a function, because of this iterative process appropriate initial parameter values need to be provided to direct the search process. This requirement can hinder the algorithms ability to converge on a global optimal solution, particularly in complex fitness landscapes, resulting in deficiently fitted models. One approach to over come the limitations of local optimizers to fit NLS is to use an algorithm from the global optimization family. Evolutionary algorithms have emerged as a powerful and robust choice to optimize non-linear models [5]. In this manuscript we will explore the use of an evolutionary algorithm and a Quasi-Newton based algorithm to fit the parameters of a common dose-response model used in the field of sport and exercise science.

It has been cited that one of the limitations of the dose-response model we will examine is its high sensitivity to the initial parameter values it is given during the fitting process [4]. However, this apparent behaviour maybe due to the use of predominately local based optimizers to fit the model and not due to any inherent bias in the model structure itself. Having the ability to fit robust and accurate dose-response models would be considered extremely advantageous in the field of exercise science as these models could be used to plan and prescribe optimal doses of exercise activities to elicit desired physiological responses with a high degree of confidence. This would allow coaches, scientists and health practitioners to guide the development of their athletes and patients with a higher degree of safety and control. Thus the ability of various optimization algorithms to improve the fit, robustness and predictive accuracy of this non-linear model presents an interesting problem from both theoretical and applied aspects.

The rest of this paper is organized as follows: Section 2 will cover the background to the problem in more detail and introduce the structure of the non-linear model, along with the parameters that are required to be fit using a NLS optimization based process. We will also present a background on the algorithms that will be used to perform the optimizations and the rationale for choosing them. Section 3 will detail the experimental design, how the models will be fit and the methods used to compare the performance of the different optimization algorithms. Section 4 will present the results of the experiments. Section 5 will discuss the results of the experiments in more detail. Finally, Section 6 will present the conclusions and possible future work.

## 2. Background

### 2.1. Physical Exercise Dose-Response Model

The Banister Impulse Response model, was developed to overcome the difficulty of translating the results of controlled research studies into practice. While a large range of studies had explored the relationship between a physical exercise training intervention and the subsequent response to that training intervention, no formal model existed that mapped objective training data to a prediction of the resulting response. To address the need for more clarity and control in the planning process Banister et al, developed the Impulse Response model (IR) [2]. The model asserts the response to a single exercise training dose has both a positive and negative physiological response. One response induces positive physiological adaptations typically referred to as a "fitness" and the other is a negative response referred to as a "fatigue". This is represented mathematically in the following manner:

$$p_t = p_0 + K_1 \sum_{i=1}^{t-1} w^i e^{-\frac{n}{r1}} + K_2 \sum_{i=1}^{t-1} w^i e^{-\frac{n-i}{r2}} \qquad (1)$$

Where $p_0$ is an initial baseline level, $K1$ and $K2$ being the positive weighting factors for fitness and fatigue, these values represent the rate at which an individual can recover from a training stimulus, $r^1$ and $r^2$ represent the time decay until fitness and fatigue return to baseline, and finally $w^i$ is the measured exercise training workload stimulus from a single session. The aforementioned parameters of the model are calibrated by minimizing the difference between the model predictions and the observed responses. Figure 1 illustrates an example model fit (grey line) and the observed responses (black dots) over a 166 day period.

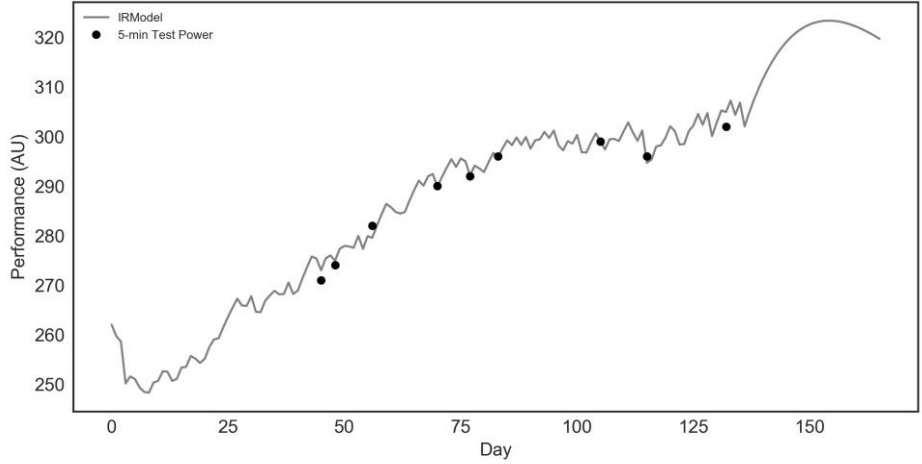

Figure 1.: IR Model Predictions - Observed Test Score

## *2.2. Parameter Optimization*

In order to fit the parameters of the IR model and ensure that it returns robust and accurate predictions, the parameters are fit by minimizing the differences between the predicted and observed responses using non-linear least squares regression. In order to perform the NLS minimization we compare the performance of a local and global optimization algorithm.

### *2.2.1. Limited Memory Broyden–Fletcher–Goldfarb–Shanno Algorithm*

The Broyden–Fletcher–Goldfarb–Shanno algorithm (BFGS) is a type of hill climbing algorithm which seeks to find the minima of a twice continuously differentiable cost function by iteratively improving an initial parameter estimation. The Limited Memory Broyden–Fletcher–Goldfarb–Shanno (L-BFGS) algorithm extends the approach of the BFGS algorithm by using a more memory friendly approach to compute its updates at each iteration [7]. Both algorithms are classed as Quasi-Newton methods because they require only the gradient of the cost function at each iteration which allows them to converge at a rate which is superlinear [12]. For these reasons variants of BFGS algorithms have been a popular choice for large scale parameter optimization [9] and NLS problems [6, 11].

### *2.2.2. Differential Evolution Algorithm*

Differential Evolution (DE) is a population based search algorithm which seeks to find solutions to a problem by manipulating a candidate population of solutions with evolutionary inspired operations such as crossover and mutation [1]. DE overcomes some of the limitations of traditional local search algorithms such as the need for a differentiable cost function and the ability to utilize parallel computing procedures. DE achieves these qualities by employing a direct search producer were by it looks for points close to its current position which achieves a lower cost function value, all in the absence of gradient information. DE combines the concept of direct search with parameter mutation, combination and selection type operations to generate new parameters solutions and direct the search procedure in a robust self-organizing fashion [13]. To date DE has shown to be a promising choice for the optimization of nonlinear dose-response modeling [8].

## 3. Experimental Design

In this experiment we focus on comparing the performance of two optimization algorithms applied to the task of fitting a nonlinear dose-response model when given a varied set of initial parameter values. In order to test the performance of both algorithms we arrange the experiment as follows : Firstly we generate a set of initial random parameter values by sampling from a constant distribution using the intervals values displayed in Table 1.

| Intervals | $p_0$ | $K_1$ | $K_2$ | $r^1$ | $r^2$ |
| --- | --- | --- | --- | --- | --- |
| Upper | 240 | 0.01 | 0.01 | 1 | 1 |
| Lower | 340 | 10 | 10 | 100 | 100 |

Table 1.

The Banister IR model is then fit in a NLS fashion using the previously described algorithms, and the randomly generated initial parameters, to find the minimum value of the cost function. Once the algorithm has converged, or has reached its maximal allowable cost function evaluations the optimized parameter values are extracted, along with the optimized model fit, loss score calculated on a hold out data set calculated as the sum of the squared error between predicted and observed values and optimization process duration. This process was repeated 1000 times with a new set of random initial parameter values for each experimental run.

In order to fit the IR model an open source data set was used, this data-set consisted of 166 exercise training sessions which are used as inputs to the IR model and real world observed responses which form a criterion measure in the fitting process. The data set is available at [3].

### 3.0.1. Algorithm Variants & Settings

In this experiment we compare a version of the L-BFGS algorithm to a seeded and unseed version of DE algorithm, the rational for this is to perform a closer comparison of the algorithms when the search process is started from a similar area of the search space. And also from a randomly initialized area as in the unseeded DE algorithm. In order to generate a seeded population of values for the DE Seeded option the initial randomized parameter values, which are provided to the L-BFGS algorithm, are used to generate a Gaussian distribution from which 19 random seeded values are selected in addition to the initial parameter values themselves. The generated normal distribution had a mean value equal to its random initial parameter and a standard deviation equal to ± 5% of that parameter value. The purpose of generating the distribution in such a way was to provide the DE algorithm with enough diversity to create new solution candidates to explore the search space but also to focus the area from which it starts to explore.

The following settings for the L-BFGS algorithm were used in all experimental runs, Maximum function evaluations: 15000, Maximum iterations: 15000, Maximum variable metric corrections: 10, convergence tolerance: 1e-11. The settings used for both DE algorithms (Seeded/ Unseeded) are as follows, Maximum iterations: 1000, population size: 20, convergence tolerance: 1e-11, mutation: (0.5, 1), recombination: 0.7. Bounds are set on all parameters using the following upper and lower values: (0, 1000), (0.01, 1000), (0.01, 1000), (1, 120), (1, 120). All operations are carried out using the scipy.optimize module as per the python3.6 Anaconda distribution.

### 3.0.2. Statistical Analysis

Descriptive statistics are used to describe the mean, 95% confidence intervals (CI) and standard deviation (Std Dev) of the parameter values over 1000 experimental runs. A one-way ANOVA was used to test the null hypothesis that two or more algorithms have the same mean performance measured by the model fit R-squared scores and loss scores on a holdout data set over the 1000 experimental runs.

## 4. Results

The mean, 95% confidence intervals and standard deviation over the 1000 experimental runs are presented in Table 2 for each parameter value and algorithm.

The results of the one-way ANOVA test on the fitted R-Squared scores are: *F statistic* = 264.71, *p* = 0.000. The results for the ANOVA test on the hold out data set mean loss scores are: *F statistic* = 90.07, *p* = 0.000. Figures 2 and 3 display the mean and 95% CI for the fitted R-squared scores and hold out losses over the 1000 experimental runs for each algorithm.

Figure 4 displays the distribution of total process time for each algorithm over the 1000 experimental runs.

| Algorithm | | p0 | K1 | K2 | r1 | r2 |
|---|---|---|---|---|---|---|
| L-BFGS | Mean | 290 | 6.04 | 5.7 | 47.8 | 49.5 |
| | CI Lower | 243 | 0.1 | 0.1 | 1 | 1 |
| | CI Upper | 338 | 21.7 | 17.37 | 96.7 | 98.4 |
| | Std Dev | 30 | 6.5 | 5.4 | 30.3 | 30.7 |
| DE Seeded | Mean | 229 | 2.7 | 2.7 | 41.3 | 36.6 |
| | CI Lower | 217 | 0.0 | 0.0 | 12.0 | 10.0 |
| | CI Upper | 291 | 9.2 | 9.2 | 85.3 | 50.0 |
| | Std Dev | 23 | 2.9 | 2.9 | 18.6 | 15.4 |
| DE Random | Mean | 224 | 77.0 | 77.0 | 29.1 | 25.9 |
| | CI Lower | 221 | 0.1 | 0.1 | 26.0 | 10.0 |
| | CI Upper | 225 | 724.6 | 724.6 | 48.6 | 30.0 |
| | Std Dev | 1 | 181.7 | 181.7 | 5.2 | 4.7 |

Table 2.: Mean, 95% CI and Std Dev of the optimal model parameters over 1000 experimental runs.

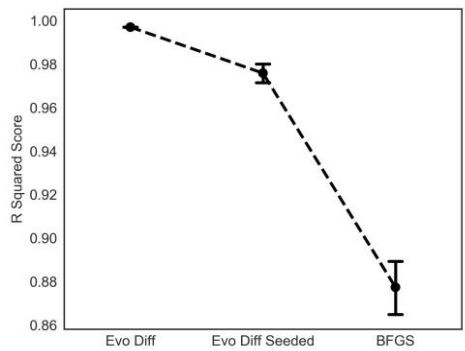

Figure 2.: Mean and 95% CI of the fitted R-squared scores

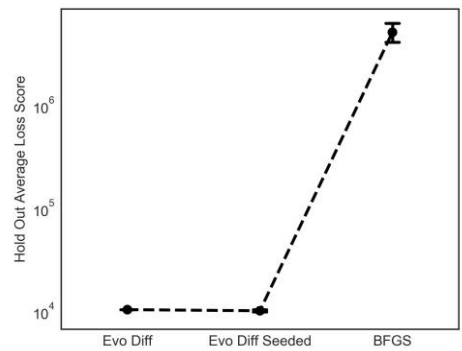

Figure 3.: Mean and 95% CI of the hold out set losses

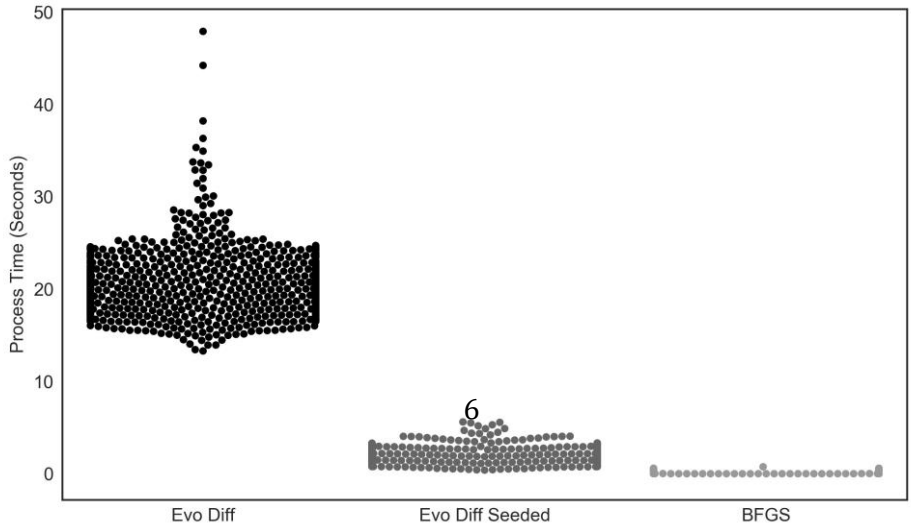

Figure 4.: Process Running Time

## 5. Discussion

This experiment set out to test the robustness and performance of a local and global optimization algorithm to fit the parameters of a non-linear dose response model when initialized with a random set of values. Robustness and performance measures were collected over 1000 experimental runs for all variants of the algorithms. We observed from the results that the global optimization algorithms DE and DE seeded had the lowest variance across all parameters compared to the L-BFGS local optimizer. Table 2 shows the standard deviation of the parameter values, the DE random algorithm demonstrated lower variability across the $p_0$, $r^1$ and $r^2$ parameters but higher variability in the $K_1$ and $K_2$ parameters. This is likely due to the larger search space that these parameters span, it is reasonable to postulate that the higher variance is the result of the DE random algorithm exploring more of this space as it attempts to converge on a global optimum. Conversely due to the limited ability to fully explore the larger search space the DE Seeded and L-BFGS optimizers demonstrate less variability.

When we begin to compare the optimizers on measures of model based robustness and performance using the optimal parameter sets across each experimental run, we can observe that the optimizers display statistically significant differences. Visual comparisons of the mean and 95% confidence intervals in figures 2 and 3, show the ability of the DE random and DE seed algorithms to consistently achieve a strong model fit and low error when predicting response values on a holdout data set. In comparison to both DE variants the L-BFGS optimizer displayed higher variance and a worse performance across the experimental runs. In the case of predicting the responses on a holdout data set the L-BFGS was particularly poor. The reason for this could be contributed to the inherent higher sensitivity of L-BFGS algorithm to initial parameter settings. While a reasonably informed initial parameter set may result in an acceptable model fit and hold out prediction performance, in this study it appears that the L-BFGS optimizer is unable to effectively overcome and deal with unsuitable initial parameter values generated from a random constant distribution with a reasonable span in this experiment.

One surprising result from this experiment is that the DE seeded optimizer dose comparably well across both performance measures when given only a small set of initial random parameters to seed it's population with. The DE seeded algorithm shows strong robustness to different initial parameter sets and tends to repeatedly converge on a globally optimal search space as demonstrated by the low standard deviation in parameter values (Table 2) and high performance when fitting the model and predicting responses on a holdout data set. This high performance is also coupled with a faster process time compared to the DE random algorithm and is only slightly slower than the L-BFGS optimizer as shown in Figure 4. This is likely attributed to a more exploitative search of the solution space due to the lower diversity in the seeded population compared to the randomly initialized population used in the DE random optimizer.

## 6. Conclusions & Future Work

In this paper we set out to compare the performance and robustness of a local versus global optimization algorithm when fitting the parameters of a non-linear dose response model. Based on our observations we found that the differential evolution global optimizer is capable of repeatedly finding optimal model parameter values demonstrated by strong model fits and performance on a holdout data set. The local search L-BFGS optimizer displayed a low robustness to varying initial parameter settings and had a poor performance when predicting values on a holdout data set. A seeded version of the differential evolution algorithm displayed the best overall performance across all measures. These result would suggest that when fitting the Banister IR model a differential evolution algorithm seeded with reasonable parameters may provide a fast, robust and high performing alternative to local search algorithms such as L-BFGS. Future work should be conducted to confirm these findings using a wider set of algorithm setting and variants, as well as larger more diverse real world data sets.